\documentclass{article} 
\usepackage[T1]{fontenc}
\usepackage{preprint_style,times}

\usepackage{amsmath,amsfonts,bm}

\def\eqref#1{equation~\ref{#1}}

\def\1{\bm{1}}

\DeclareMathAlphabet{\mathsfit}{\encodingdefault}{\sfdefault}{m}{sl}
\SetMathAlphabet{\mathsfit}{bold}{\encodingdefault}{\sfdefault}{bx}{n}

\usepackage{booktabs,tabularx,multirow}
\usepackage{algorithm}
\usepackage{algpseudocode}
\usepackage{graphicx}
\usepackage{wrapfig}
\usepackage{needspace}
\usepackage{xcolor}
\usepackage{hyperref}
\usepackage{url}

\title{VideoEvoHarness: Self Evolving Agent Harness from Unlabeled Videos for Video Understanding}

\author{%
\parbox[t]{\textwidth}{\raggedright
\textbf{Weixin Xu\(^{1,2,6}\), Yifan Liu\(^{3}\), Yi Wang\(^{4}\), Mingfei Rong\(^{5}\), Zhenyu Yang\(^{1}\), Zhisheng Zhong\(^{6,\dagger}\), Shengsheng Qian\(^{1,2,*}\), Changsheng Xu\(^{1,2}\)}\\
\normalfont
\(^{1}\)Institute of Automation, Chinese Academy of Sciences, \(^{2}\)School of Advanced Interdisciplinary Sciences, University of Chinese Academy of Sciences, \(^{3}\)School of Computer Science and Communication Engineering, Jiangsu University of Science and Technology, \(^{4}\)AICOMO, \(^{5}\)College of Computer Science, Inner Mongolia University, \(^{6}\)Tencent Hunyuan\\
\texttt{xuweixin2025@ia.ac.cn}}}

\begin{document}

\maketitle
{\renewcommand{\thefootnote}{\fnsymbol{footnote}}%
\footnotetext[1]{Corresponding author: shengsheng.qian@nlpr.ia.ac.cn}%
\footnotetext[2]{Project leader.}}

\begin{abstract}
Video understanding agents rely on a harness to acquire evidence, use tools, and reason, but improving it requires substantial effort. Recent self evolution methods seek to optimize harnesses through execution feedback to reduce human intervention. However, existing methods often rely on annotated video questions and answers or focus on updating model parameters, leaving harness evolution from unlabeled videos underexplored. VideoEvoHarness jointly evolves a Questioner and a Solver harness from unlabeled videos with frozen model parameters. The Questioner constructs questions and computes reference answers directly from video memory records. Question quality assessments and Solver execution trajectories guide question generation policy revisions. The Solver supports evolving skills, strategies, hyperparameters, and tools using execution feedback. Executable strategies define workflows in code, reducing repeated interpretation of natural language instructions. Each round, candidate harnesses are compared on identical validation questions, with validated improvements retained for later use. Experiments on benchmarks and a dedicated benchmark question set, using videos disjoint from evolution videos, show higher overall accuracy than initially and transfer gains across backbones. Overall accuracy on this set also improves progressively across rounds. This supports our foundational framework for harness evolution from unlabeled videos, opening a promising direction for self evolving video agents. Code is available at \url{https://github.com/Wei-Xin-Xu/VideoEvoHarness}.
\end{abstract}
 
\section{Introduction}
\label{sec:introduction}

Recent advances in large vision language models have improved their ability to understand video content and reason about events as they unfold over time \citep{bai2025qwen25vl,he2024malmm,ren2024timechat,shu2025videoxl,diko2025rewind}. These advances have enabled the emergence of video agents. They use tools to retrieve evidence and select information for reasoning and answering questions \citep{wang2024videoagent,fan2024videoagent,long2026m3agent,ma2025drvideo}. Their performance therefore depends not only on the model but also on how effectively tools are used. These patterns of agent behavior are coordinated by the harness, which integrates instructions, reusable skills, tools, and execution procedures around the model \citep{yao2023react,cai2024latm,khattab2024dspy}.

Designing an agent harness requires decisions across its components and their interactions. When these decisions remain fixed, newly observed failure trajectories require manual revision of the relevant harness components. This motivates agent self evolution, which enables an agent to analyze its own execution outcomes and automatically evaluate and revise its harness \citep{zhang2026selfharness}. Existing work on general agent systems illustrates this process. GEPA refines prompts by reflecting on execution traces \citep{agrawal2026gepa}. Meta Agent Search uses a model to write agent programs, evaluate them, and develop new versions from previous designs and their results \citep{hu2025adas}. These developments motivate exploring self evolution for video agents, allowing task experience to inform subsequent evidence acquisition and reasoning.

\begin{figure}[t]
\centering
\includegraphics[width=\linewidth]{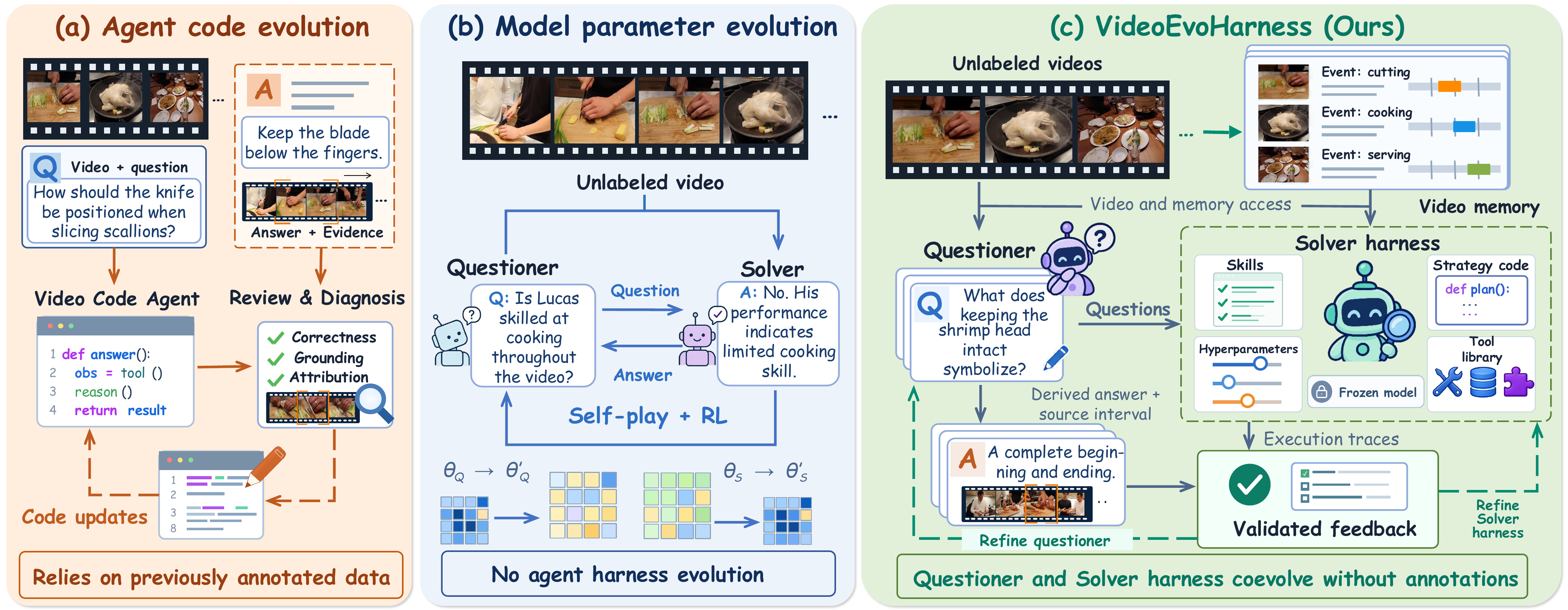}
\caption{Comparison of video agent evolution paradigms. (a) Agent code evolution revises agent code through review and diagnosis, and relies on previously annotated questions, answers, and evidence. (b) Model parameter evolution builds supervision from unlabeled video through Questioner and Solver interaction, but updates model parameters and leaves the agent harness unchanged. (c) VideoEvoHarness derives questions, reference answers, and source intervals from video memory, and uses validated feedback to support Questioner refinement and Solver updates across skills, strategy code, hyperparameters, and tools, while the model stays frozen.}
\label{fig:overview}
\end{figure}

Applying agent evolution to video understanding requires reliable evaluation, yet constructing video questions and verified answers is costly and demands careful quality control. The annotation process must examine video content, interpret temporal context, and verify answers against visual evidence \citep{fu2025videomme}. In video agent evolution, one line of work depends heavily on previously annotated questions, answers, and evidence, and modifies the harness by evolving agent code, as illustrated in Figure~\ref{fig:overview}(a) \citep{cui2026metavideoagent}. Raw videos, by contrast, are abundant online and typically come without accompanying questions or answers. This gap motivates us to consider using unlabeled videos directly as a source of evolution feedback. To do so, an agent must construct evaluation tasks from unlabeled video. As illustrated in Figure~\ref{fig:overview}(b), EvoVid introduces a Questioner and a Solver to derive supervision from raw video clips and iteratively update model parameters \citep{huang2026evovid}. However, sampled clips may omit parts of an event and its temporal context, making it difficult to generate informative questions. This leads to our first challenge: \emph{How can video agents derive effective evolution feedback from unlabeled videos without externally supplied questions and answers?}

Beyond obtaining feedback, we must also consider how to evolve the agent harness. As shown in Figure~\ref{fig:overview}(b), EvoVid updates only model parameters without evolving the agent harness \citep{huang2026evovid}. MetaVideoAgent evolves the harness, but limits its updates to agent code, as shown in Figure~\ref{fig:overview}(a) \citep{cui2026metavideoagent}. Another approach under exploration adapts agent behavior by revising prompts and reusable textual skills \citep{agrawal2026gepa,tu2026visualclaw}. In agent environments, these skills can build on established tools for search, file manipulation, and program execution. Video tasks, however, depend on temporal localization and the analysis of visual evidence. Their toolchains remain under active development, with different systems providing different combinations of these capabilities \citep{wang2024videoagent,fan2024videoagent}. Textual skill refinement alone may therefore be insufficient for complex video tasks. Executing these skills also requires the model to interpret instructions and translate them into tool calls. Variations in this interpretation can make an intended strategy difficult to reproduce consistently. This leads to our second challenge: \emph{How can a video agent evolution framework be designed to evolve overall execution strategies rather than rely solely on textual skill optimization?}

To address these challenges, we introduce \textbf{VideoEvoHarness}, a framework in which a Questioner and a Solver evolve together from unlabeled video, as shown in Figure~\ref{fig:overview}(c). The Questioner generates supervision for improvement, while the Solver modifies its harness under that supervision. Task construction thus adapts alongside task execution, allowing unlabeled videos to supply both evaluation tasks and directions for improvement. The underlying model parameters remain fixed throughout the process.

For the first challenge, \textbf{Questioner evolution} adapts the supervision constructed from structured visual memory. The Questioner generates questions about events and entities recorded in local video segments and computes reference answers through operations on the memory. The associated video segments also provide a basis for analyzing the causes of Solver failures. We assess five aspects of question quality: answer accuracy without video evidence, answer accuracy with supporting memory, question type coverage, difficulty for the current Solver, and the proportion of questions removed by quality filtering. Solver execution trajectories provide additional feedback. The model uses this information to revise the question generation policy, allowing the supervision process to evolve with experience from successive task executions.

For the second challenge, \textbf{Solver harness evolution} supports updates to textual skills, executable strategy code, hyperparameters, and tools. Textual skills refine reasoning instructions, while strategy code specifies how evidence is retrieved and tools are composed. Writing these procedures as code makes their control flow explicit and reduces reliance on repeated model interpretation of natural language instructions. Hyperparameter updates adapt observation and evidence allocation, while the tool operation strategy expands the set of available operations or modifies existing tool code. The Solver applies a sampling strategy to short videos and uses memory to locate relevant frames in long videos, without being given the Questioner's source event. The model then uses execution feedback to propose changes and evaluates the revised harness on subsequent tasks. Validated improvements are retained for later iterations. Together, these mechanisms allow the agent to revise both what it can do and how it organizes those capabilities.

Our contributions are summarized as follows:
\begin{itemize}
\item We introduce \textbf{VideoEvoHarness}, a framework for evolving video agent harnesses from unlabeled videos without external question and answer annotations or updates to model parameters.
\item We develop a feedback mechanism for \textbf{Questioner evolution}. Question quality assessment and Solver execution trajectories inform revisions to the generation policy, adapting the supervision derived from unlabeled video.
\item We develop a mechanism supporting \textbf{Solver harness evolution} across textual skills, executable strategy code, hyperparameters, and tools. Candidate updates target reasoning and execution procedures, with executable strategies specifying how frames are sampled, memory is retrieved, and existing tools are used.
\item Evaluations on standard video understanding benchmarks and a dedicated benchmark question set, using videos disjoint from evolution videos, show higher overall accuracy than the initial harness, progressive gains on the dedicated set across rounds, and transfer gains across backbones, supporting harness evolution.
\end{itemize}

\section{Related Work}
\label{related_work}

\paragraph{Video Understanding Agents.}Video understanding agents extend Video-LLMs with active observation, structured memory, and role coordination. Observation-planning agents decide what to inspect: VideoAgent uses an LLM controller to iteratively identify and compile question-relevant information, with vision-language models serving as visual retrieval and perception tools \citep{wang2024videoagent}. VideoZoomer learns coarse-to-fine temporal focusing, TimeSearch-R interleaves adaptive search with completeness self-verification, and LensWalk lets the reasoner select each observation's temporal range and sampling density \citep{ding2026videozoomer,pan2026timesearchr,li2026lenswalk}. Memory-structured agents organize video content for targeted retrieval. Fan et al. store event descriptions and object-centric tracks in memory accessed through segment-localization and object-query tools, while VideoTree and HOMER construct query-adaptive or hierarchical representations for multi-stage retrieval \citep{fan2024videoagent,wang2025videotree,ji2026homer}. Role-coordination agents modularize inference; VideoMind orchestrates planner, grounder, verifier, and answerer roles \citep{liu2026videomind}. Although these methods adapt evidence acquisition to each query, their prompts, tools, strategies, and hyperparameters generally remain fixed after design or training. \textbf{VideoEvoHarness} instead makes the video agent harness evolvable.

\paragraph{Self Evolving Agents.}Self evolving agents adapt tasks, instructions, or executable systems from interaction feedback \citep{agrawal2026gepa,hu2025adas,fernando2024promptbreeder}. WebRL evolves curricula from failed web trajectories \citep{qi2025webrl}; GEPA converts trajectory reflection into prompt updates, while automated agent design searches over executable agent programs \citep{agrawal2026gepa,hu2025adas,fernando2024promptbreeder}. Vision-Zero extends label-free multi-agent self-play to visual tasks \citep{wang2026visionzero}. In video, data- and model-evolution methods improve training supervision: Video-STaR cycles between instruction generation and fine-tuning using source labels, while ReWatch-R1 synthesizes video-grounded reasoning traces for post-training \citep{zohar2025videostar,zhang2026rewatchr1}. Video-Zero and EvoVid instead coevolve Questioner and Solver policies from raw videos without human annotations, but primarily update Video-LLM parameters rather than the runtime harness \citep{zhang2026videozero,huang2026evovid}. Harness-oriented methods are closer to our setting: VisualClaw accumulates deployment failures into memory and skill banks, whereas MetaVideoAgent evolves modular agent code but requires supplied questions, gold answers, and annotated evidence intervals for failure diagnosis \citep{tu2026visualclaw,cui2026metavideoagent}. \textbf{VideoEvoHarness} instead enables task construction and the execution harness to coevolve from unlabeled videos without human-authored questions, answers, or evidence annotations, while keeping model parameters frozen.

\section{Method}
\label{sec:method}

We introduce \textbf{VideoEvoHarness}, a framework for evolving video agents through tasks derived from unlabeled video. Its central idea is to improve both the questions that reveal an agent's limitations and the harness that determines how it addresses them. A Questioner refines task construction from video observations, while a Solver evolves its evidence acquisition and reasoning procedures. We first describe their interaction, then present the evolution mechanism for each component.

\subsection{Framework Overview}
\label{sec:framework}

\begin{figure}[t]
\centering
\includegraphics[width=\linewidth]{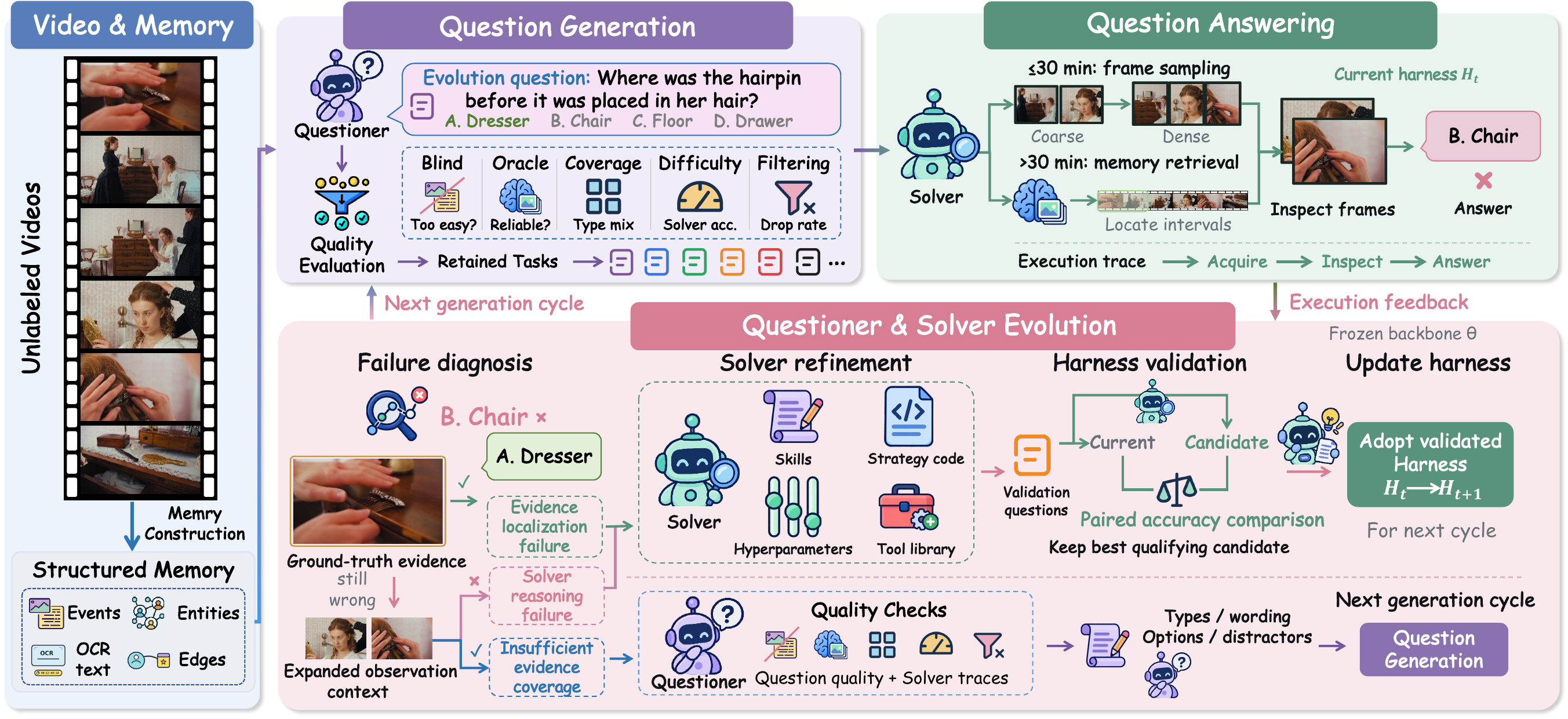}
\caption{Overview of the VideoEvoHarness framework. The Questioner constructs and filters tasks from structured video memory, while the Solver acquires visual evidence through frame sampling or memory retrieval. Execution feedback supports revisions to the question generation policy and the Solver's skills, strategy code, hyperparameters, and tool operations. Validated updates enable both agents to evolve from unlabeled videos while keeping model parameters fixed.}
\label{fig:pipeline}
\end{figure}

Given unlabeled videos \(\mathcal{V}\) and a pretrained model \(f_\theta\), we seek to improve video understanding by evolving the surrounding harness while keeping \(\theta\) fixed. The Questioner constructs evaluation tasks from the videos, and the Solver tests its harness on these tasks. Their results guide evolution. Figure~\ref{fig:pipeline} summarizes this cycle, from memory construction and question generation to answering, diagnosis, and harness updates (Algorithm~\ref{alg:videoevoharness}). We represent the system at iteration \(t\) as
\begin{equation}
\label{eq:harness_state}
\Phi_t = (\phi_t, H_t), \qquad
H_t = (P_t, C_t, \lambda_t, \mathcal{T}_t),
\end{equation}
where \(\phi_t\) specifies the Questioner's generation policy and \(H_t\) denotes the Solver harness. Its components are textual skills and prompts \(P_t\), executable strategies \(C_t\), hyperparameters \(\lambda_t\), and a tool library \(\mathcal{T}_t\). Together, they determine what evidence the Solver observes and how it uses that evidence to answer a question.

We build on Qwen-MM-Plugins \citep{qwenmmplugins2026} for video reading and memory construction (Appendix~\ref{app:memory_construction}). The Questioner derives questions, reference answers, and source segments from hierarchical memory containing events, entities, and visible text. The Solver uses two evidence acquisition paths according to video duration. For short videos, it uses a single overview or chunk-wise coarse sampling followed by targeted detail inspection. Long videos are searched through memory to locate relevant segments before inspecting their original frames. Hyperparameters control temporal windows, sampling density, and resolution, as detailed in Appendix~\ref{app:evidence_acquisition}. The reference answer and designated source segment are withheld from the Solver. Question evaluation and Solver execution trajectories inform updates to task generation and the Solver harness.

\subsection{Questioner Evolution}
\label{sec:questioner}

\paragraph{Question generation.}
The central challenge in learning from unlabeled video is to derive an evaluation signal from observations alone. We address this by generating questions whose reference answers can be computed from structured video records. The visual memory associates event summaries, entities, actions, and visible text with temporal intervals. The Questioner selects a source segment and constructs a question supported by its records. Temporal questions use event timestamps. Counting questions derive reference answers by counting entities of a given type in the memory of the selected segment. Text questions retrieve recorded visual content. The resulting reference is derived from the memory rather than inferred from the Solver's response. Each question also retains its source segment, providing a reference for analyzing evidence retrieval. A video therefore supplies both an answer evaluation signal and a basis for diagnosing how the answer was obtained.

\paragraph{Question quality evaluation.}
The questions generated from visual memory should reveal gaps in the Solver's video understanding. However, some can be answered from their wording or answer options without consulting the video. We measure accuracy without video evidence using a Blind probe before filtering and exclude questions it answers correctly from the Solver's task set. An Oracle probe measures accuracy with supporting memory; questions it answers incorrectly are also excluded. We assess question type coverage against target proportions specified in the question generation configuration. We estimate difficulty from Solver accuracy within each question type and favor types with intermediate accuracy. Finally, we track the proportion of generated questions removed by filtering, since excessive filtering reduces the number available for evolution. We combine these five criteria into a question quality score:
\begin{equation}
\label{eq:questioner_quality}
\mathcal{J}_{Q}
= -w_{\mathrm{b}} L_{\mathrm{b}} + w_{\mathrm{o}} A_{\mathrm{o}}
- w_{\mathrm{c}} C + w_{\mathrm{d}} D - w_{\mathrm{r}} R,
\end{equation}
where \(L_{\mathrm{b}}\) is normalized accuracy above chance without video evidence, and \(A_{\mathrm{o}}\) is the fraction of questions answered correctly by the Oracle with supporting memory. \(C\) is the distance between the retained and target question type distributions. \(D\) is the fraction of retained questions whose type has intermediate Solver accuracy. \(R\) is the discarded fraction above the level expected from random Blind guesses. The weights balance these criteria. Detailed definitions are given in Appendix~\ref{app:questioner_details}.

\paragraph{Generation policy refinement.}
We provide the model with the question evaluation results, Solver execution trajectories, and the current generation policy. The model identifies weaknesses in question generation and proposes revisions to \(\phi_t\). These revisions act on question types, question forms, answer options, and distractor selection. For example, frequent success without visual evidence motivates replacing descriptive cues with relations between observations or drawing distractors from similar temporal contexts. Solver performance provides additional context about which task types are already mastered and which remain difficult. The revised policy changes how subsequent questions are constructed. This makes question generation an adaptive source of supervision rather than a fixed preprocessing step.

\subsection{Video Agent Harness Evolution}
\label{sec:harness_evolution}

\paragraph{Failure analysis.}
Questioner evolution revises how tasks are constructed, while Solver harness evolution revises how they are solved. The Solver obtains evidence by sampling frames under its strategy or by retrieving memory. A wrong answer alone does not indicate whether the relevant evidence was missed or incorrectly used. We compare evidence conditions as appropriate for each path, using source segments retained during question construction. We preserve each path's evidence representation during these comparisons. A correct answer after supplying source evidence suggests an evidence localization failure by the Solver. A correct answer only after expanding evidence coverage suggests insufficient evidence coverage for the Questioner's question and reference answer. If the Solver still answers incorrectly under both conditions, we examine possible errors in its reasoning. We combine these comparisons with execution trajectories to identify limitations in evidence acquisition and reasoning.

\paragraph{Harness refinement.}
Failure analysis supports updates to any Solver harness component. Textual skills in \(P_t\) can shape task interpretation and reasoning through reusable prompts, such as instructions to verify entity counts against relevant frames. For executable strategies in \(C_t\), the model can write and revise code for frame sampling, evidence retrieval, tool composition, and answer formation. Hyperparameter updates in \(\lambda_t\) can control retrieval breadth, sampling density, resolution, and evidence allocation. Tool updates can extend \(\mathcal{T}_t\) with reusable operations or revise existing code. New or modified tools must pass verification and candidate harness validation before being retained.

Given the current harness and evaluation feedback, the model proposes changes to these components, including revised strategy and tool code. We apply the updated harness to subsequent tasks and use the evaluation results to refine it further. Combining these four forms of adaptation enables the Solver to adjust both the operations available for reasoning and the procedures that organize their use.

\paragraph{Improvement validation.}
The Solver tests each candidate harness on a shared validation subset of the Questioner's questions. We retain the candidate with the greatest validated improvement in accuracy. If no candidate is accepted, we preserve the current harness. Details are provided in Appendix~\ref{app:validation}. Questions and reference answers remain fixed during this comparison. Questioner revisions take effect in a later generation cycle, separating changes in task difficulty from changes in Solver capability. Accepted skills, strategies, parameter settings, and tools are incorporated into the harness for subsequent iterations. Feedback from video tasks thus accumulates as reusable reasoning and execution procedures.

\section{Experiments}
\label{sec:experiments}

\subsection{Experimental Setup}
\label{sec:setup}

\paragraph{Datasets and metrics.}
We evolve the harness on a pool of 130 videos sampled from six sources: the MLVU development split \citep{zhou2024mlvu}, long videos from Vript, VideoMarathon \citep{lin2025videomarathon}, long TVQA episodes \citep{lei2018tvqa}, ScaleLong \citep{ma2025scalelong}, and the Video-Holmes training split \citep{cheng2025videoholmes}. We use only the raw videos and discard all accompanying questions, answers, and temporal annotations. Source counts and the video duration distribution are provided in Appendix~\ref{app:evolution_videos}. These videos are disjoint from the evaluation videos, which we verify by video identifier and duration. We evaluate on three benchmarks: LongVideoBench \citep{wu2024longvideobench}, MMVU \citep{zhao2025mmvu}, and Video-Holmes \citep{cheng2025videoholmes}. All evaluations use video frames without subtitles and report multiple-choice accuracy; for MMVU, we use its multiple-choice subset. Table~\ref{tab:longvideobench} reports LongVideoBench results by question category under L1-Perception and L2-Relation, together with overall accuracy. Table~\ref{tab:mmvu} reports MMVU results across its four official disciplines and overall, alongside overall accuracy on Video-Holmes. Self generated questions support evolution and question quality analysis. Benchmark evaluations are conducted independently and do not guide evolution.

\paragraph{Baselines.}
Our primary comparison is the same frozen model under the initial harness before evolution, which separates the effect of evolution from the capability of the model. This harness follows the recommended configuration of the underlying toolkit \citep{qwenmmplugins2026} rather than a weakened setting. We compare Initial harness with VideoEvoHarness after the first and third evolution rounds, denoted R1 and R3, respectively. We evaluate two backbones: Qwen3.7 Plus \citep{qwen37plus} and Gemini 3.8 Flash \citep{gemini38flash}.

\paragraph{Implementation details.}
We build the harness on Qwen-MM-Plugins \citep{qwenmmplugins2026} and keep all model parameters frozen. Videos of at least 30 minutes are searched through memory to locate relevant segments before their frames are inspected. Videos under 30 minutes use a single overview or segment-wise coarse sampling with targeted detail inspection. Evolution proceeds for three rounds. Each round draws 30 videos from this pool and generates about 600 usable questions: 200 for evolution and 400 for validating whether proposed modifications should be accepted. The initial and evolved harnesses start from the same initial observation budget. Remaining configurations are given in Appendix~\ref{app:implementation}.

\begin{table*}[t]
\centering
\caption{Per-category results on LongVideoBench with frozen backbones. Question categories follow the official taxonomy and are grouped into L1-Perception and L2-Relation. Avg. reports accuracy over all questions within each group, and Overall reports accuracy over all questions. Scores are accuracy in percent, and the best result for each backbone is in bold.}
\label{tab:longvideobench}
\vspace{\baselineskip}
\setlength{\tabcolsep}{3pt}
\renewcommand{\arraystretch}{1.15}
\resizebox{\linewidth}{!}{%
\begin{tabular}{@{}l | ccccccccc | cccccccccc | c@{}}
\toprule
\multicolumn{1}{c|}{\multirow{2}{*}{\textbf{Method}}}
 & \multicolumn{9}{c|}{\textbf{L1-Perception}}
 & \multicolumn{10}{c|}{\textbf{L2-Relation}}
 & \multirow{2}{*}{\textbf{Overall}} \\
\cmidrule(lr){2-10}\cmidrule(lr){11-20}
 & S2E & S2O & S2A & E2O & O2E & T2E & T2O & T2A & \textbf{Avg.}
 & E3E & O3O & SSS & SOS & SAA & T3E & T3O & TOS & TAA & \textbf{Avg.} \\
\midrule
\multicolumn{1}{c|}{} & \multicolumn{20}{c}{\textbf{Qwen3.7 Plus}} \\
\midrule
Initial harness & 69.9 & \textbf{72.2} & 85.1 & 75.4 & 65.5 & 64.5 & 72.0 & 65.3 & 71.4 & 74.7 & 74.2 & 55.8 & 72.8 & 59.7 & 52.1 & 59.5 & 31.9 & 53.7 & 59.6 & 65.1 \\
\textbf{VideoEvoHarness} (R1) & 72.0 & 70.8 & 86.2 & 73.8 & \textbf{70.1} & 62.9 & \textbf{78.7} & 64.0 & 72.7 & 75.8 & 77.3 & \textbf{64.2} & 75.3 & 58.3 & \textbf{60.3} & \textbf{63.5} & 31.9 & 43.9 & 61.5 & 66.7 \\
\textbf{VideoEvoHarness} (R3) & \textbf{73.1} & 66.7 & \textbf{87.4} & \textbf{80.0} & 65.5 & \textbf{67.7} & 76.0 & \textbf{68.0} & \textbf{73.2} & \textbf{78.0} & \textbf{80.3} & 61.1 & \textbf{82.7} & \textbf{65.3} & 53.4 & 54.1 & \textbf{34.7} & \textbf{56.1} & \textbf{63.2} & \textbf{67.9} \\
\midrule
\multicolumn{1}{c|}{} & \multicolumn{20}{c}{\textbf{Gemini 3.8 Flash}} \\
\midrule
Initial harness & 83.9 & \textbf{77.8} & \textbf{94.3} & \textbf{83.1} & \textbf{80.5} & 75.8 & 74.7 & \textbf{78.7} & 81.5 & 79.1 & 72.7 & 63.2 & \textbf{82.7} & 75.0 & \textbf{67.1} & 64.9 & 47.2 & \textbf{67.1} & 69.0 & 74.8 \\
\textbf{VideoEvoHarness} (R1) & 83.9 & \textbf{77.8} & \textbf{94.3} & \textbf{83.1} & 79.3 & \textbf{79.0} & 80.0 & 74.7 & \textbf{81.8} & \textbf{82.4} & 77.3 & \textbf{69.5} & 79.0 & 73.6 & 65.8 & 67.6 & \textbf{54.2} & \textbf{67.1} & \textbf{71.0} & \textbf{76.0} \\
\textbf{VideoEvoHarness} (R3) & \textbf{86.0} & 75.0 & 93.1 & \textbf{83.1} & 79.3 & 74.2 & \textbf{81.3} & 74.7 & 81.3 & \textbf{82.4} & \textbf{78.8} & 65.3 & 76.5 & \textbf{79.2} & 61.6 & \textbf{71.6} & 48.6 & \textbf{67.1} & 70.3 & 75.4 \\
\bottomrule
\end{tabular}%
}

\end{table*}

\Needspace{18\baselineskip}
\subsection{Main Results}
\label{sec:main_results}

\noindent
\begin{wraptable}{r}{0.56\textwidth}
\vspace{-\dimexpr\baselineskip+\intextsep\relax}
\caption{Results on MMVU and Video-Holmes.}
\label{tab:mmvu}
\vspace{0.3\baselineskip}
\centering
\scriptsize
\setlength{\tabcolsep}{0.5pt}
\begin{tabular*}{\linewidth}{@{\extracolsep{\fill}}l | ccccc | c@{}}
\toprule
\multicolumn{1}{c|}{\multirow{2}{*}{\textbf{Method}}} & \multicolumn{5}{c|}{\textbf{MMVU}} & \multirow{2}{*}{\textbf{V-Holmes}} \\
\cmidrule(lr){2-6}
 & Sci. & Health. & Hum. & Eng. & \textbf{Overall} & \\
\midrule
\multicolumn{1}{c|}{} & \multicolumn{6}{c}{\textbf{Qwen3.7 Plus}} \\
\midrule
Initial harness & 76.5 & 79.3 & \textbf{87.0} & 80.9 & 80.0 & 56.5 \\
\textbf{VideoEvoHarness} (R1) & 78.4 & 82.6 & 85.9 & 81.9 & 81.4 & 57.1 \\
\textbf{VideoEvoHarness} (R3) & \textbf{79.8} & \textbf{85.1} & 83.7 & \textbf{82.9} & \textbf{82.4} & \textbf{58.0} \\
\midrule
\multicolumn{1}{c|}{} & \multicolumn{6}{c}{\textbf{Gemini 3.8 Flash}} \\
\midrule
Initial harness & 83.6 & \textbf{90.1} & 87.0 & 85.4 & 85.9 & 65.3 \\
\textbf{VideoEvoHarness} (R1) & \textbf{85.9} & 87.6 & 89.1 & \textbf{87.4} & 87.2 & 65.5 \\
\textbf{VideoEvoHarness} (R3) & 85.4 & 89.3 & \textbf{90.2} & 86.9 & \textbf{87.4} & \textbf{65.8} \\
\bottomrule
\end{tabular*}
\end{wraptable}
\noindent\textbf{Evolution from unlabeled videos.}\quad
Tables~\ref{tab:longvideobench} and~\ref{tab:mmvu} report results on three benchmarks with two frozen backbones. VideoEvoHarness (R3) improves overall accuracy over the initial harness on every benchmark: Qwen3.7 Plus gains 2.8, 2.4, and 1.5 points on LongVideoBench, MMVU, and Video-Holmes, respectively; Gemini 3.8 Flash gains 0.6, 1.5, and 0.5 points. On LongVideoBench, Qwen's L2-Relation score improves by 3.6 points, compared with 1.8 for L1-Perception. Gemini similarly gains 1.3 points on L2 despite a 0.2-point decline on L1, suggesting larger benefits for relational questions. Table~\ref{tab:mmvu} groups MMVU into Science (Sci.), Healthcare (Health.), Humanities and Social Sciences (Hum.), and Engineering (Eng.), and reports Video-Holmes (V-Holmes) overall. MMVU also exhibits discipline-specific gains: Qwen improves most in Healthcare (5.8 points), while Gemini benefits most in Humanities and Social Sciences (3.2 points). These gains are obtained using tasks generated from unlabeled videos, without updating model parameters.

\paragraph{Development across evolution rounds.}
The first round already improves overall accuracy on all three benchmarks for both backbones. Qwen3.7 Plus achieves higher overall accuracy at R3 than at R1, with LongVideoBench rising from 66.7 to 67.9, MMVU from 81.4 to 82.4, and Video-Holmes from 57.1 to 58.0. This indicates additional gains beyond the first update. Gemini 3.8 Flash shows some fluctuation: LongVideoBench drops from 76.0 at R1 to 75.4 at R3, but remains above its initial 74.8. Meanwhile, MMVU and Video-Holmes improve from 87.2 to 87.4 and 65.5 to 65.8, respectively. Thus, although later rounds do not improve every benchmark monotonically, R3 retains gains over the initial harness on all three. Because the evolution videos are disjoint from all evaluation videos, these results support transfer of the evolved harness beyond its self generated tasks.

\subsection{Evolution Effectiveness}
\label{sec:evolution_effectiveness}

\begin{figure}[t]
\centering
\setlength{\abovecaptionskip}{4pt}
\begin{minipage}[t]{0.48\textwidth}
\centering
\includegraphics[width=\linewidth]{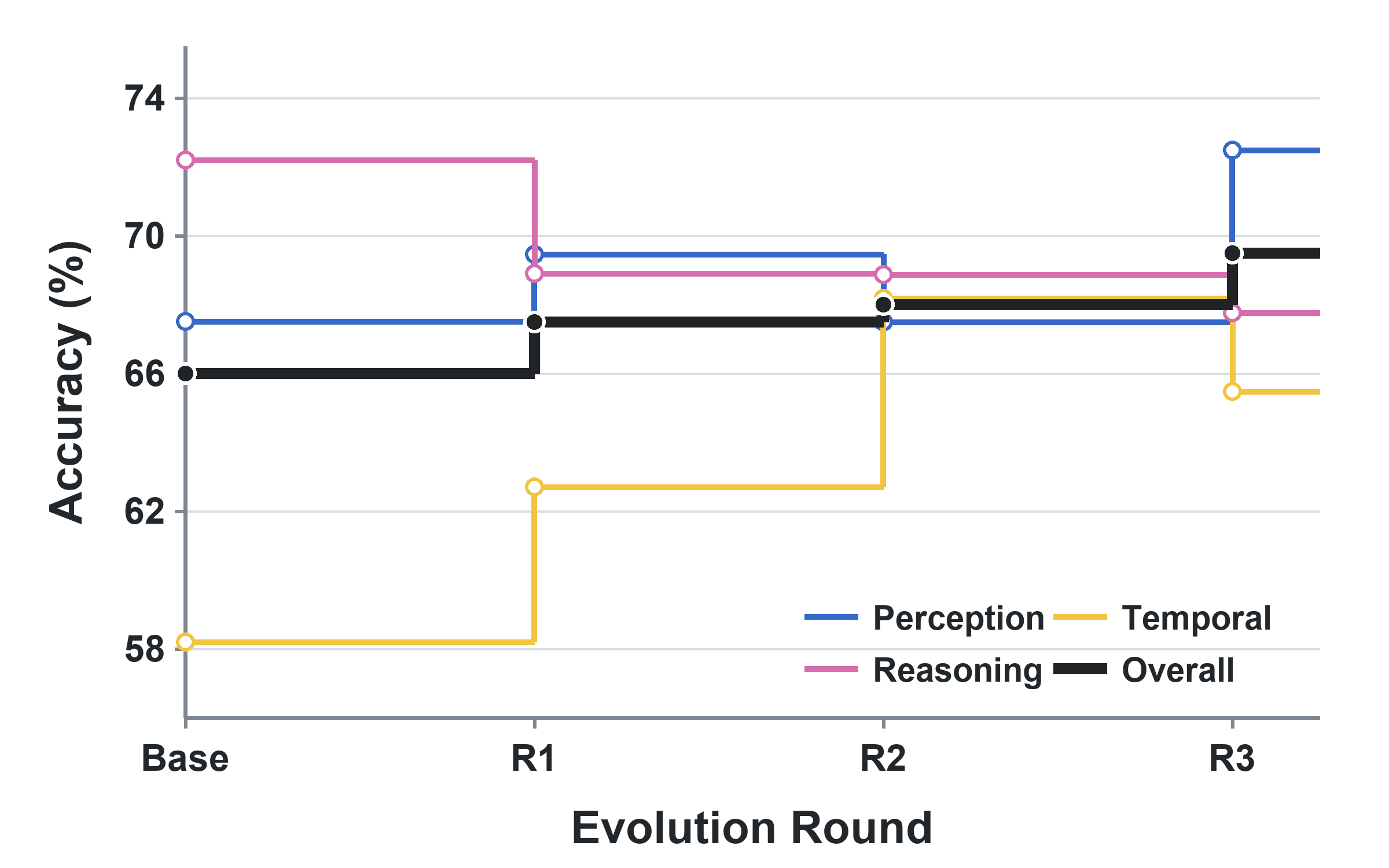}
\caption{Accuracy on the dedicated benchmark question set across evolution rounds. Base denotes the initial harness.}
\label{fig:benchmark_evolution}
\end{minipage}\hfill
\begin{minipage}[t]{0.48\textwidth}
\centering
\includegraphics[width=\linewidth]{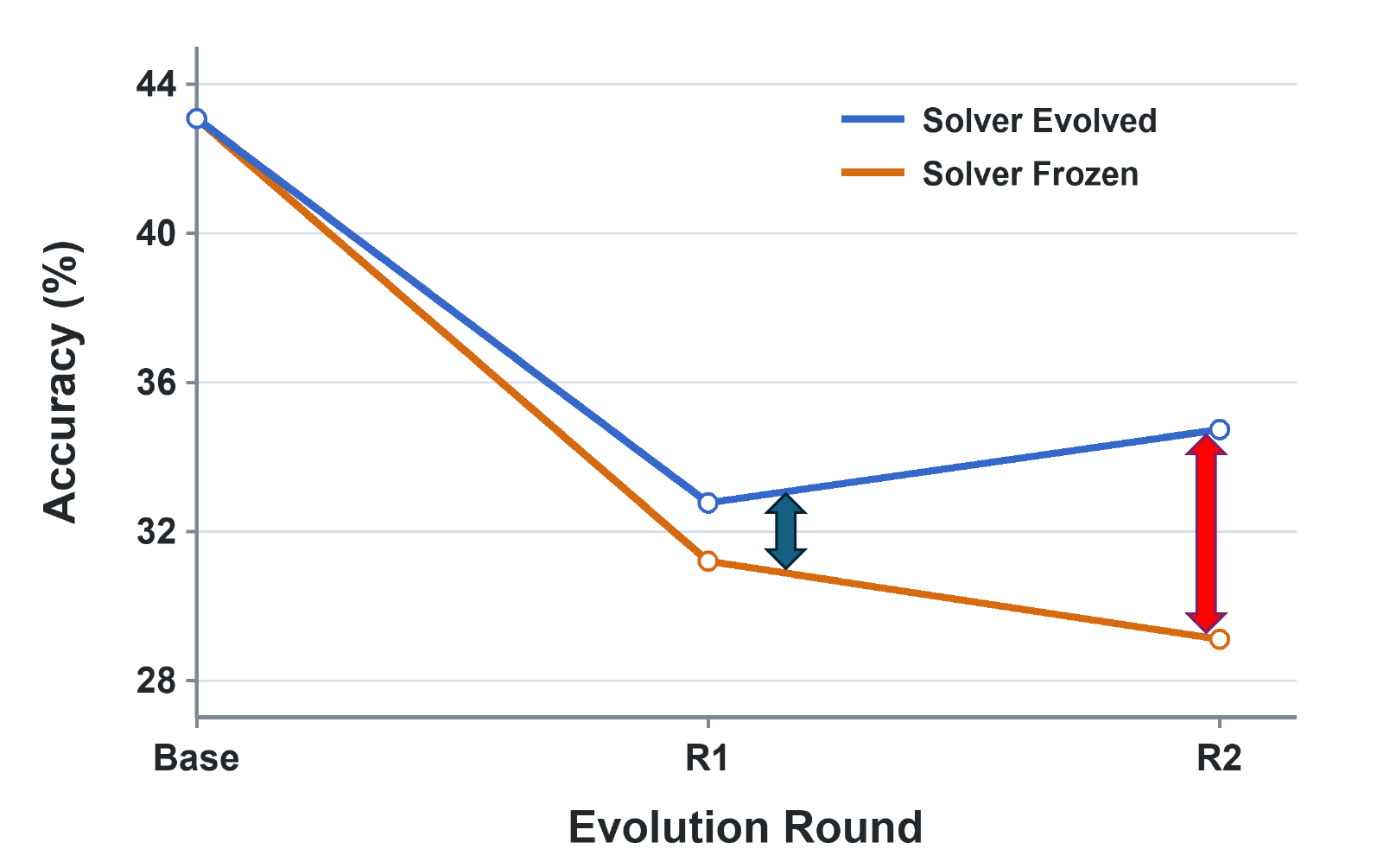}
\caption{Accuracy on generated questions across evolution rounds for the evolving and initial Solver harnesses.}
\label{fig:solver_adaptation}
\end{minipage}
\end{figure}

\noindent\textbf{Performance across evolution rounds.}\quad
For a balanced assessment, we evaluate the initial harness (Base) and each round's saved harness with Qwen3.7 Plus on the same dedicated benchmark question set of 400 questions, grouped into perception, temporal grounding, and reasoning (Appendix~\ref{app:evaluation_set}). These questions do not guide evolution. Figure~\ref{fig:benchmark_evolution} shows overall accuracy increasing across evolution rounds, indicating that later updates provide benefits beyond the first round. Perception improves overall despite intermediate fluctuations, while temporal grounding rises substantially before declining in the final round, remaining above its initial level. These trends show that the evolved harness improves both visual perception and temporal grounding, although progress within each group is not monotonic. In contrast, reasoning declines, mainly driven by lower performance on Topic and knowledge questions. This suggests that current evolution primarily improves fine-grained evidence localization. This provides a direction for subsequent evolution: strengthening global information integration and summarization while preserving these gains.

\noindent\textbf{Question difficulty and Solver adaptation.}\quad
Using Qwen3.7 Plus, we compare the evolving Solver harness with its initial version on questions generated by the evolving Questioner at each round. The Questioner is evolved in the first two of the three evolution rounds. Both Solver versions answer the same questions within a round and are scored against identical references. Figure~\ref{fig:solver_adaptation} shows both declining from Base to R1; at R2, the evolving Solver recovers while the frozen Solver continues to decline. The continued decline of the unchanged Solver harness indicates that later question sets are more challenging, with a growing proportion of questions probing unresolved capability gaps in the initial harness. Meanwhile, the evolving Solver outperforms the frozen version at both R1 and R2, with a larger advantage at R2. Despite fluctuations in absolute accuracy, this widening gap indicates that the evolved Solver adapts more effectively to the changing question distribution. Together with the gains on the dedicated set in Figure~\ref{fig:benchmark_evolution}, these results support the effectiveness of coevolving task generation and execution.

\subsection{Ablation Studies}
\label{sec:ablation}

On the same dedicated benchmark question set, we examine cross-backbone harness transfer, filtering and validation, and Solver component evolution. Perc., Temp., and Reas. report accuracy for perception, temporal grounding, and reasoning and synthesis, respectively; All reports overall accuracy. Scores are percentages. Unless otherwise stated, we use Qwen3.7 Plus.

\begin{table}[t]
\centering
\begin{minipage}[t]{0.48\textwidth}
\centering
\caption{\strut Transfer of the R3 harness from Gemini 3.8 Flash to Qwen3.7 Plus.\strut}
\label{tab:transfer_ablation}
\vspace{0.3\baselineskip}
\footnotesize
\setlength{\tabcolsep}{1pt}
\begin{tabular*}{\linewidth}{@{\extracolsep{\fill}}l*{4}{c}@{}}
\toprule
\textbf{Harness} & \textbf{Perc.} & \textbf{Temp.} & \textbf{Reas.} & \textbf{All} \\
\midrule
Initial & 67.5 & 58.2 & \textbf{72.2} & 66.0 \\
Transferred & 68.5 & 64.5 & 70.0 & 67.8 \\
Own evolved & \textbf{72.5} & \textbf{65.5} & 67.8 & \textbf{69.5} \\
\bottomrule
\end{tabular*}
\end{minipage}\hfill
\begin{minipage}[t]{0.48\textwidth}
\centering
\caption{\strut Ablation of question filtering and candidate validation over three evolution rounds.\strut}
\label{tab:loop_ablation}
\vspace{0.3\baselineskip}
\footnotesize
\setlength{\tabcolsep}{1pt}
\begin{tabular*}{\linewidth}{@{\extracolsep{\fill}}l*{5}{c}@{}}
\toprule
\textbf{Variant} & \textbf{Perc.} & \textbf{Temp.} & \textbf{Reas.} & \textbf{All} & \textbf{N} \\
\midrule
\textbf{Full framework} & \textbf{72.5} & \textbf{65.5} & 67.8 & \textbf{69.5} & 3 \\
\hspace{0.4em}No validation & 68.0 & 61.8 & 68.9 & 66.5 & \textbf{39} \\
\hspace{0.4em}No filtering & 66.0 & 64.5 & \textbf{71.1} & 66.8 & 1 \\
\bottomrule
\end{tabular*}
\end{minipage}
\end{table}

\noindent\textbf{Transfer across backbones.}\quad
Table~\ref{tab:transfer_ablation} evaluates the transfer of the harness evolved for three rounds with Gemini 3.8 Flash to Qwen3.7 Plus. Initial uses Qwen's original harness; Transferred reuses Gemini's R3 harness without further evolution; Own evolved uses the harness evolved with Qwen itself. Transferred raises overall accuracy from 66.0 to 67.8, showing that an evolved harness can benefit another backbone without target-specific evolution. Temporal grounding gains 6.3 points, while reasoning drops by 2.2 points, indicating that transfer benefits differ across capabilities. Own evolved further improves overall accuracy to 69.5, exceeding Transferred by 1.7 points. Thus, direct reuse provides cross-backbone gains, while evolving the harness with the target model itself yields the best overall performance.

\noindent\textbf{Effectiveness of filtering and validation.}\quad
Table~\ref{tab:loop_ablation} examines question filtering and candidate validation. N counts harness updates retained over three evolution rounds. Full framework retains three updates and achieves the highest overall accuracy (69.5). No validation retains all 39 proposed updates without checking their benefit, yet reaches only 66.5 overall accuracy, with temporal grounding falling from 65.5 to 61.8. This contrast shows that update quantity alone does not ensure progress; update quality and effectiveness matter more. Without question filtering, low-quality tasks weaken the feedback used to propose changes, and most candidates fail to improve validation accuracy. Only one strategy update is retained, yielding 66.8 overall accuracy. These results highlight complementary roles: filtering provides informative tasks to guide useful proposals, while validation selects updates with measured gains. Together, they support more effective evolution within the three-round budget.

\noindent
\begin{wraptable}{r}{0.46\textwidth}
\vspace{-\dimexpr\baselineskip+\intextsep\relax}
\caption{Ablation of Solver evolution. Column maxima are in bold.}
\label{tab:component_ablation}
\vspace{0.3\baselineskip}
\centering
\footnotesize
\setlength{\tabcolsep}{1pt}
\begin{tabular*}{\linewidth}{@{\extracolsep{\fill}}l*{4}{c}@{}}
\toprule
\textbf{Harness} & \textbf{Perc.} & \textbf{Temp.} & \textbf{Reas.} & \textbf{All} \\
\midrule
\textbf{Full harness} & \textbf{72.5} & \textbf{65.5} & 67.8 & \textbf{69.5} \\
\hspace{0.4em}Strategy & 68.0 & 62.7 & 67.8 & 66.5 \\
\hspace{0.4em}Hyperparams. & 69.5 & 60.0 & 66.7 & 66.3 \\
Initial harness & 67.5 & 58.2 & \textbf{72.2} & 66.0 \\
\bottomrule
\end{tabular*}
\end{wraptable}
\noindent\textbf{Effectiveness of component evolution.}\quad
Table~\ref{tab:component_ablation} evaluates strategy and hyperparameter evolution. Initial harness denotes the reference configuration, and Full harness corresponds to Own evolved in Table~\ref{tab:transfer_ablation}. Strategy and Hyperparams. retain only the corresponding component changes. Skills and tools remain unchanged within the current three-round budget and are therefore omitted. Relative to Initial harness, Strategy improves perception and temporal grounding by 0.5 and 4.5 points, respectively, highlighting the effectiveness of execution logic for coordinating temporal evidence. Hyperparams. gains 2.0 and 1.8 points in these categories, demonstrating the contribution of sampling and retrieval adjustments to visual evidence acquisition. Combining both updates yields complementary benefits: Full harness reaches 69.5 overall accuracy, exceeding Strategy and Hyperparams. by 3.0 and 3.2 points, respectively. It gains 3.5 points overall over Initial harness, with improvements of 5.0 in perception and 7.3 in temporal grounding, while Initial harness retains higher reasoning accuracy (72.2 versus 67.8). The combined configuration thus achieves the strongest overall performance and the highest perception and temporal grounding scores, bringing together the benefits of execution logic and evidence allocation. These results demonstrate the effectiveness of component evolution in improving Solver performance.

\section{Conclusion}
\label{sec:conclusion}

VideoEvoHarness coevolves a Questioner and a Solver harness from unlabeled videos with frozen model parameters. The Questioner generates tasks from video memory and refines its generation policy using question quality assessments and Solver execution trajectories. Validated execution feedback supports updates to the Solver's textual skills, executable strategies, hyperparameters, and tools. Executable strategies encode reusable workflows, reducing repeated interpretation of natural language instructions. Within each round, candidates are compared on identical validation questions, with accepted improvements retained for subsequent tasks. Experiments on three benchmarks with two backbones and on a dedicated benchmark question set show higher overall accuracy than the initial harness and transfer gains across backbones. Overall accuracy on this set also improves progressively across rounds. These results highlight harness evolution from unlabeled videos as a promising direction for adaptable video understanding agents.

\bibliography{references}
\bibliographystyle{preprint_style}

\clearpage
\appendix
\section*{ Appendix}

\section{Implementation Details}
\label{app:implementation}

\subsection{Video Memory Construction}
\label{app:memory_construction}

We construct video memory using the hierarchical graph pipeline provided by Qwen-MM-Plugins \citep{qwenmmplugins2026}. First, scene boundaries are detected from frame differences in hue, lightness, and saturation, using frames sampled at 0.25 fps. Short segments are merged and long segments are split using minimum and maximum duration settings of 30 and 300 seconds. These memory segments are distinct from the fixed 300-second chunks used for Solver observation in Appendix~\ref{app:evidence_acquisition}. In the frame-input construction path, each segment is sampled at 0.5 fps with a cap of 250 frames before being sent to the visual model. These construction settings are separate from the Solver's frame inspection budget.

For each segment, the model produces a local graph containing events, entities, visible text, and their relations. Event records include actions and time intervals; entity records include types, attributes, and descriptions; visible-text records retain the text and its time interval. Segment-relative timestamps are converted to positions in the full video, and each record retains its parent segment identifier. A text-based aggregation stage groups segments into higher-level events and constructs a video-level description. Node descriptions are embedded using Qwen3-Embedding-8B \citep{zhang2025qwen3} for semantic retrieval, while the graph retains identifiers and temporal links for direct lookup.

The saved graph and retrieval index are reused across questions. The Questioner computes references by accessing the recorded fields, counts, and timestamps, and retains the corresponding source segments for diagnosis. The long-video Solver instead retrieves memory using the question and inspects the associated video intervals without being told which segments were used to construct the question. Memory construction is a toolkit component on which our evolution framework builds, rather than a new representation learned by the evolution loop. Table~\ref{tab:hyperparameters} summarizes the construction defaults and the initial harness settings.

\subsection{Question Quality Estimation}
\label{app:questioner_details}

Let \(\mathcal{G}\) denote the generated questions before filtering and \(\mathcal{D}\) the retained subset. Blind accuracy \(b\) is measured on \(\mathcal{G}\). For questions with \(K\) answer options, the excess Blind accuracy and excess rejection rate used in Equation~\ref{eq:questioner_quality} are
\begin{equation}
\label{eq:excess_blind}
L_{\mathrm{b}} = \max\!\left\{0,\frac{b-1/K}{1-1/K}\right\}, \qquad
R = \max\!\left\{0,1-\frac{|\mathcal{D}|}{|\mathcal{G}|}-\frac{1}{K}\right\}.
\end{equation}
Both terms account for correct Blind guesses at chance. Retained questions have failed the Blind probe, so their Blind accuracy is not used to estimate generation quality. Oracle accuracy \(A_{\mathrm{o}}\) is the fraction of evaluated questions answered correctly with supporting memory. Questions answered incorrectly by this probe are excluded from the Solver's task set.

We use the total variation distance between the retained question type distribution and the target proportions as the coverage penalty \(C\). Target proportions are specified within the question generation configuration rather than estimated from benchmark statistics. Difficulty \(D\) is the fraction of retained questions whose type has Solver accuracy within a prescribed interval, with default bounds of 0.3 and 0.7. These accuracies are estimated from the tasks used for diagnosis. The quality profile supplies feedback for generation policy proposals. Questioner revisions take effect when tasks are generated again, rather than when the Solver is rerun on an unchanged task set.

\subsection{Solver Candidate Validation}
\label{app:validation}

In each round, about 600 usable generated questions are divided into an evolution subset of 200 questions and a validation subset of 400 questions. The evolution subset supports failure diagnosis and candidate proposals, while Solver comparisons use the same questions and references from the validation subset. For \(N\) validation questions, let \(a_i\) be the reference answer and let \(\hat a_i\) and \(\hat a_i'\) be the current and candidate answers. We define the paired correctness difference and its mean as
\begin{equation}
\label{eq:candidate_gain}
d_i = \mathbf{1}\{\hat a_i'=a_i\}-\mathbf{1}\{\hat a_i=a_i\}, \qquad
\widehat{\Delta} = \frac{1}{N}\sum_{i=1}^{N} d_i,
\end{equation}
where \(\mathbf{1}\{\cdot\}\) is the indicator function. A candidate passes the accuracy criterion when \(\widehat{\Delta}>0\). Among candidates passing this criterion and the resource checks, the one with the highest validation accuracy is retained. The current harness is preserved if none qualifies. This generated validation subset is distinct from the dedicated benchmark question set in Appendix~\ref{app:evaluation_set}.

\subsection{Generated Tool Verification}
\label{app:tool_verification}

The model proposes each tool as a Python function, together with its name, input signature, purpose, supported question types, and test cases. Each test specifies input arguments and an expected output or condition. For example, an entity-counting tool accepts a time interval and an entity type, retrieves a matching memory segment, and counts its entity entries. A supplied test can request \texttt{PERSON} entries between 10 and 20 seconds and require a nonnegative numeric result.

We first inspect the code's syntax tree for disallowed imports, calls, and attribute accesses. We then execute the supplied tests against a selected video's memory through the toolkit. Tests check equality to a supplied value, a specified minimum, a nonempty result, or an output type. All supplied assertions must pass. Errors and failed assertions are returned to the model for revision, with up to three generation attempts.

These tests are supplied by the same model and check basic execution and output properties rather than independently establishing correctness. Passing them makes the tool available during candidate Solver evaluation, not part of the retained harness. The tool is added to the persistent library only if its candidate harness is selected under the validation procedure in Appendix~\ref{app:validation}.

\subsection{Video Evidence Acquisition}
\label{app:evidence_acquisition}

Evidence acquisition depends on video duration. We describe the initial harness configuration and the sampling defaults below; parameters exposed to evolution can change in later rounds. The Questioner uses memory to construct tasks on all paths, but reference answers and designated source intervals are withheld from the Solver during ordinary execution.

\paragraph{Sampling path.}
Videos of at most 5 minutes receive a single overview at a requested rate of 1 frame per second (fps), followed by one answer call. Videos longer than 5 minutes but shorter than 30 minutes are divided into consecutive 300-second chunks. Each chunk is inspected at 1 fps to produce question-relevant observations and candidate intervals for closer inspection. Candidate intervals receive 5 seconds of context on each side; overlapping intervals or those separated by at most 2 seconds are merged. Up to six intervals, each capped at 60 seconds, are inspected at 2 fps. Their textual observations are then combined in one final answer call. Overview and detail inspection use the toolkit's \texttt{normal} and \texttt{large} visual budgets, respectively \citep{qwenmmplugins2026}. The default limit is 192 frames per visual model call, not per question; frame and payload limits can reduce the effective sampling rate.

\paragraph{Memory path.}
Videos of at least 30 minutes use memory to locate intervals before inspecting their frames. Initially, segment-level and node-level searches each request up to six results. Their parent segments are merged and deduplicated, and up to two segments supply the initial text evidence within a 6,000-character budget. The Solver can then use memory tools for up to three rounds and select up to three intervals for frame inspection. The initial configuration adds 24 seconds of context on each side of a matched node's time interval, subject to a 60-second window cap. Frames are requested at 1 fps, with a shared budget of 192 frames across the selected intervals. These are starting values rather than fixed settings for all evolved harnesses.

\paragraph{Failure analysis.}
We assess evidence coverage using execution trajectories and, where applicable, additional observations. These may include denser frame sampling or memory records from source segments and broader video context. Diagnostic comparisons preserve the evidence representation of each path. Together, the evidence comparisons and trajectories help identify possible localization, coverage, and reasoning limitations.

\subsection{Hyperparameters}
\label{app:hyperparameters}

Table~\ref{tab:hyperparameters} summarizes key hyperparameters, including those for memory construction, question generation, evidence acquisition, and model calls. For evolvable parameters, we report the initial values, which may change across rounds. Frame rates are requested sampling rates, subject to frame-count and payload limits. Output-token limits apply to individual model calls, not entire question-solving runs.

\begin{table}[htbp]
\centering
\caption{Key hyperparameters. Sampling settings for memory construction and Solver execution are listed separately. Model-call entries specify temperature / maximum output tokens.}
\label{tab:hyperparameters}
\small
\setlength{\tabcolsep}{5pt}
\renewcommand{\arraystretch}{1.04}
\begin{tabularx}{\linewidth}{@{}Xl@{}}
\toprule
Parameter & Value \\
\midrule
\multicolumn{2}{@{}l}{\textbf{Memory construction}} \\
Scene detection rate & 0.25 fps \\
Minimum / maximum segment duration & 30 / 300 s \\
Subgraph sampling rate / frame cap per segment & 0.5 fps / 250 \\
\midrule
\multicolumn{2}{@{}l}{\textbf{Question generation and quality}} \\
Number of answer options \(K\) & 4 \\
Configured weights \((w_{\mathrm b},w_{\mathrm o},w_{\mathrm c},w_{\mathrm d},w_{\mathrm r})\) & \((2,1,1,1,0.5)\) \\
Intermediate type-accuracy interval & \([0.3,0.7]\) \\
\midrule
\multicolumn{2}{@{}l}{\textbf{Solver evidence acquisition}} \\
Direct / memory routing cutoffs & 300 / 1,800 s \\
Coarse observation chunk duration & 300 s \\
Overview / detail sampling rates & 1 / 2 fps \\
Maximum detail intervals / interval duration & 6 / 60 s \\
Detail padding per side / merging gap & 5 / 2 s \\
Sampling-path frame cap per visual call & 192 \\
Memory search results per entry / retained segments & 6 / 2 \\
Initial memory-text budget & 6,000 characters \\
Memory-tool rounds / selected replay intervals & 3 / 3 \\
Replay sampling rate / shared frame cap & 1 fps / 192 \\
Replay padding per side / window cap & 24 / 60 s \\
\midrule
\multicolumn{2}{@{}l}{\textbf{Model calls and tool generation}} \\
Solver answering & 0.1 / 2,048 \\
Memory-tool decisions & 0.0 / 3,000 \\
Hyperparameter proposals & 0.3 / 1,200 \\
Tool-routing policy proposals & 0.4 / 1,600 \\
Memory-strategy code generation & 0.4 / 2,600 \\
Memory-tool code generation & 0.3 / 2,000 \\
Maximum tool-generation attempts & 3 \\
\bottomrule
\end{tabularx}
\end{table}

\subsection{Dedicated Benchmark Question Set}
\label{app:evaluation_set}

The dedicated benchmark question set contains 400 questions and is used for Figure~\ref{fig:benchmark_evolution} and Section~\ref{sec:ablation}. We first define ten question types, then sample from each benchmark according to the types it covers, so that no single benchmark determines the overall composition. Each question retains its original answer options and reference answer, and evaluation follows the protocol of Section~\ref{sec:setup}. The set remains fixed across rounds and analyses; missing answers count as incorrect. For reporting, the ten types are grouped into perception, temporal grounding, and reasoning and synthesis, which contain 200, 110, and 90 questions. Analyses report accuracy within each group and over the full set, since an individual benchmark contributes too few questions for a separate comparison. Table~\ref{tab:evaluation_set} reports the composition together with this grouping.

\begin{table}[t]
\centering
\caption{Composition of the dedicated benchmark question set used in Figure~\ref{fig:benchmark_evolution} and Section~\ref{sec:ablation}. Question types are divided into three reporting groups, with the group size given in parentheses. Each column reports the number of questions contributed by one benchmark, and Share is the proportion of the full set. LongVB, V-Holmes, and V-MME denote LongVideoBench, Video-Holmes, and Video-MME.}
\label{tab:evaluation_set}
\vspace{\baselineskip}
\footnotesize
\begin{tabular*}{\linewidth}{@{\extracolsep{\fill}}ll@{}*{8}{c@{}}}
\toprule
\textbf{Group} & \textbf{Question type} & \textbf{LVBench} & \textbf{MLVU} & \textbf{LongVB} & \textbf{V-Holmes} & \textbf{MMVU} & \textbf{V-MME} & \textbf{Total} & \textbf{Share} \\
\midrule
\multirow{4}{*}{\shortstack[l]{\textit{Perception}\\(200)}}
 & Counting & 3 & 8 & 2 & 0 & 1 & 6 & 20 & 5.00 \\
 & Entities and attributes & 15 & 13 & 25 & 4 & 6 & 17 & 80 & 20.00 \\
 & Actions and events & 10 & 15 & 12 & 6 & 5 & 12 & 60 & 15.00 \\
 & Spatial context & 7 & 6 & 13 & 3 & 4 & 7 & 40 & 10.00 \\
\midrule
\multirow{2}{*}{\shortstack[l]{\textit{Temporal}\\(110)}}
 & Temporal order & 10 & 15 & 14 & 7 & 5 & 9 & 60 & 15.00 \\
 & Retrieval and localization & 10 & 12 & 8 & 5 & 3 & 12 & 50 & 12.50 \\
\midrule
\multirow{4}{*}{\shortstack[l]{\textit{Reasoning}\\(90)}}
 & Causal and multi step & 7 & 7 & 4 & 14 & 10 & 3 & 45 & 11.25 \\
 & Topic and knowledge & 3 & 6 & 0 & 5 & 9 & 2 & 25 & 6.25 \\
 & Anomaly and egocentric & 1 & 6 & 0 & 1 & 1 & 1 & 10 & 2.50 \\
 & Summarization & 4 & 2 & 2 & 0 & 1 & 1 & 10 & 2.50 \\
\midrule
\multicolumn{2}{l}{\textbf{Total}} & 70 & 90 & 80 & 45 & 45 & 70 & 400 & 100.00 \\
\bottomrule
\end{tabular*}
\end{table}

\subsection{Evolution Video Collection}
\label{app:evolution_videos}

The evolution pool contains 130 videos drawn from six source collections. Tables~\ref{tab:evolution_videos} and~\ref{tab:evolution_video_durations} report their source counts and duration distribution, respectively, using the source identifiers in the video collection. Only raw videos are used; accompanying questions, answers, and temporal annotations are excluded.

\begin{table}[htbp]
\centering
\begin{minipage}[t]{0.48\linewidth}
\vspace{0pt}
\centering
\caption{Source counts of the evolution videos.}
\label{tab:evolution_videos}
\vspace{\baselineskip}
\small
\begin{tabular*}{\linewidth}{@{\extracolsep{\fill}}lr@{}}
\toprule
\textbf{Source} & \textbf{Videos} \\
\midrule
\texttt{mlvu\_dev} & 33 \\
\texttt{vript\_long} & 30 \\
\texttt{videomarathon} & 29 \\
\texttt{tvqa\_long} & 23 \\
\texttt{scalelong} & 8 \\
\texttt{video\_holmes\_train} & 7 \\
\midrule
\textbf{Total} & \textbf{130} \\
\bottomrule
\end{tabular*}
\end{minipage}\hfill
\begin{minipage}[t]{0.46\linewidth}
\vspace{0pt}
\centering
\caption{Duration distribution of the evolution videos.}
\label{tab:evolution_video_durations}
\vspace{\baselineskip}
\small
\begin{tabular*}{\linewidth}{@{\extracolsep{\fill}}lr@{}}
\toprule
\textbf{Duration (min)} & \textbf{Videos} \\
\midrule
\(d \leq 5\) & 30 \\
\(5 < d \leq 30\) & 70 \\
\(30 < d \leq 60\) & 25 \\
\(d > 60\) & 5 \\
\midrule
\textbf{Total} & \textbf{130} \\
\bottomrule
\end{tabular*}
\end{minipage}
\end{table}

\section{Overall Evolution Algorithm}
\label{app:algorithm}

Algorithm~\ref{alg:videoevoharness} summarizes the evolution loop using the notation of Equation~\ref{eq:harness_state}. Memory is constructed once and reused. Feedback \(\mathcal F_t\) combines question quality reports, Solver trajectories, and failure diagnosis. Reference answers and source labels support evaluation and diagnosis but are withheld during ordinary solving. Candidates may modify any enabled component of \((P_t,C_t,\lambda_t,\mathcal T_t)\); unchanged components are retained.

\begin{algorithm}[H]
\caption{VideoEvoHarness Evolution}
\label{alg:videoevoharness}
\small
\algrenewcommand{\textproc}{\textit}
\renewcommand{\algorithmicrequire}{\textbf{Input:}}
\renewcommand{\algorithmicensure}{\textbf{Output:}}
\begin{algorithmic}[1]
\Require Videos \(\mathcal V\), frozen model \(f_\theta\), initial policies \((\phi_0,H_0)\), rounds \(T\)
\Ensure Updated Questioner \(\phi_T\) and Solver harness \(H_T\)
\State \(\mathcal M\gets\Call{BuildMemory}{\mathcal V}\)
\State \(\mathcal B\gets\Call{GenerateAndFilter}{\phi_0,\mathcal M}\)
\State \(r\gets\mathrm{false}\)
\For{\(t=0,\ldots,T-1\)}
    \State \(H_{t+1}\gets H_t\), \(\phi_{t+1}\gets\phi_t\)
    \If{\(r\) or a scheduled question refresh is due}
        \State \(\mathcal B\gets\Call{GenerateAndFilter}{\phi_t,\mathcal M}\); \(r\gets\mathrm{false}\)
    \EndIf
    \State \((\mathcal D_t^{\mathrm{evo}},\mathcal D_t^{\mathrm{val}})\gets\Call{SelectTasks}{\mathcal B,t}\)
    \State \(\mathcal F_t\gets\Call{CollectFeedback}{H_t,\mathcal D_t^{\mathrm{evo}},\mathcal B,\mathcal M}\)
    \State \(\mathcal H_t\gets\Call{ProposeHarnesses}{H_t,\mathcal F_t}\)
    \State \(\mathcal H_t\gets\Call{Precheck}{\mathcal H_t}\)
    \State Compute \(\widehat\Delta(H')\) for each \(H'\in\mathcal H_t\) on the same \(\mathcal D_t^{\mathrm{val}}\)
    \State \(\mathcal H_t^+\gets\{H'\in\mathcal H_t:\widehat\Delta(H')>0\ \text{and}\ \Call{ResourceCheck}{H'}\}\)
    \If{\(\mathcal H_t^+\neq\emptyset\)}
        \State \(H_{t+1}\gets\operatorname*{arg\,max}_{H'\in\mathcal H_t^+}\widehat\Delta(H')\)
        \State Register any newly selected strategy or tool code
        \State \(\mathcal Q_t\gets\Call{ProposeQuestioners}{\phi_t,\mathcal F_t}\)
        \State \(J_t\gets\Call{PairedQuality}{\{\phi_t\}\cup\mathcal Q_t,\mathcal M,\mathcal F_t}\)
        \State \(\mathcal Q_t^+\gets\{\phi'\in\mathcal Q_t:J_t(\phi')>J_t(\phi_t)\}\)
        \If{\(\mathcal Q_t^+\neq\emptyset\)}
            \State \(\phi_{t+1}\gets\operatorname*{arg\,max}_{\phi'\in\mathcal Q_t^+}J_t(\phi')\)
            \State \(r\gets\mathrm{true}\)
        \EndIf
    \EndIf
\EndFor
\State \Return \((\phi_T,H_T)\)
\end{algorithmic}
\end{algorithm}

\textit{Precheck} applies code restrictions and supplied tests before candidate evaluation (Appendix~\ref{app:tool_verification}). The gain \(\widehat\Delta(H')\) is measured relative to \(H_t\) using Equation~\ref{eq:candidate_gain}, and \textit{ResourceCheck} enforces the enabled budget limits. \textit{PairedQuality} regenerates and filters questions for each policy using the same video scope, generation budget, and random seed. It returns the scores \(J_t(\phi)=\mathcal J_Q(\phi)\) from Equation~\ref{eq:questioner_quality}, with difficulty assessed using the shared feedback \(\mathcal F_t\). Accepted Questioner revisions trigger a question-bank refresh in the next round, leaving the current validation questions unchanged. Benchmark evaluations are separate from this loop.

\section{Case Study}
\label{app:case_study}

\subsection{Generated Question Examples}
\label{app:question_examples}

\begin{figure}[htbp]
\centering
\includegraphics[width=\linewidth,height=0.42\textheight,keepaspectratio]{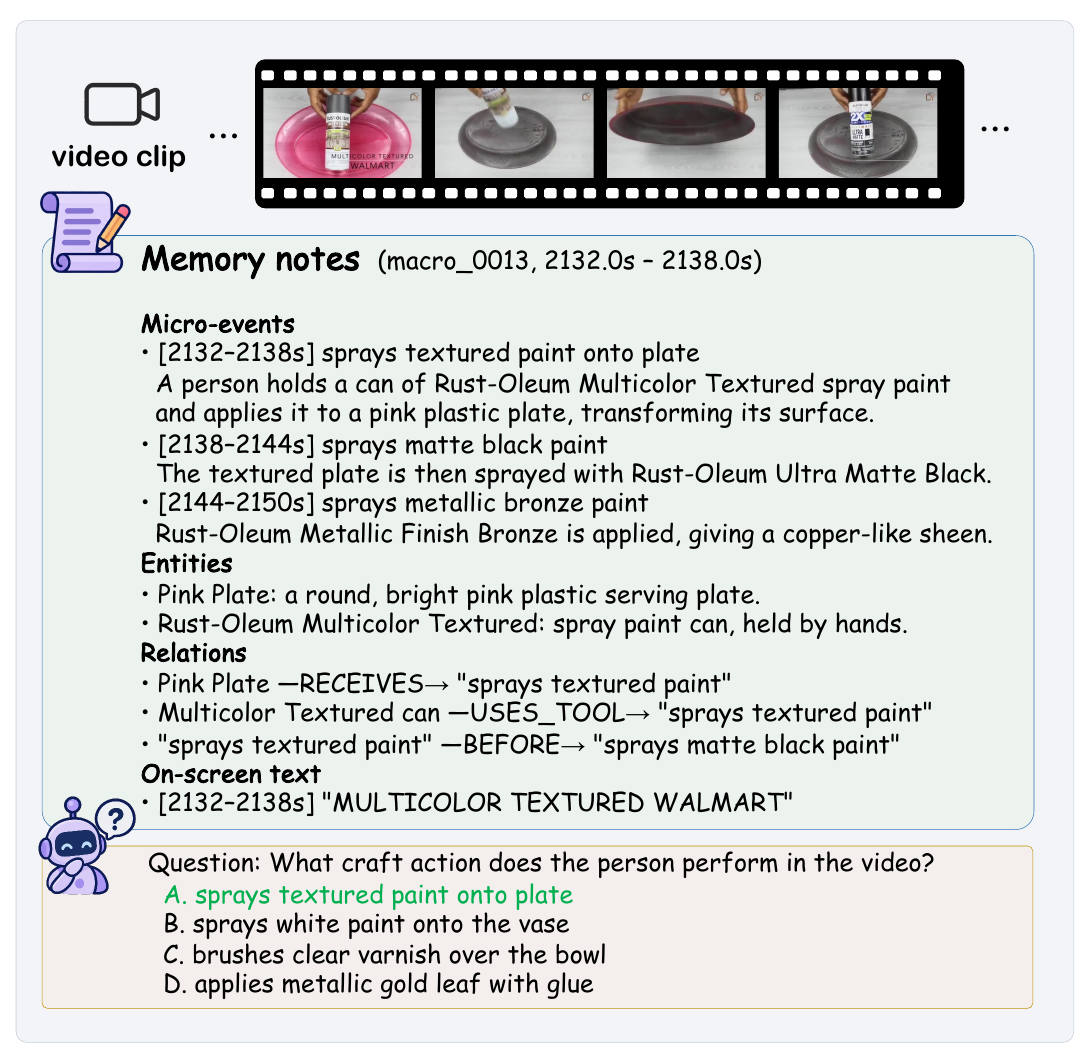}
\caption{Generated action-recognition question. The micro-event record supports answer A: spraying textured paint onto a plate.}
\label{fig:question_action}
\vspace{0.5\baselineskip}
\includegraphics[width=\linewidth,height=0.42\textheight,keepaspectratio]{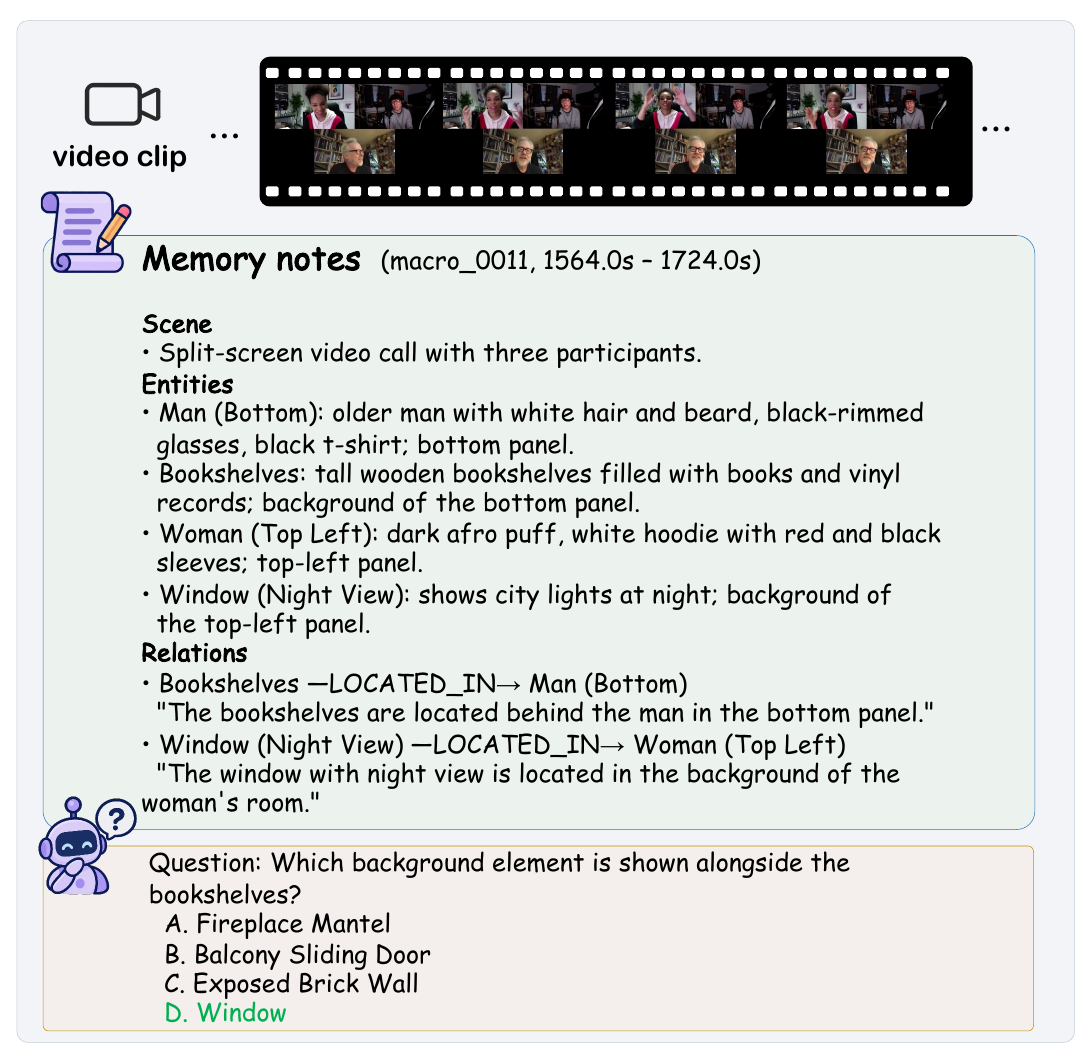}
\caption{Generated object-reasoning question based on entity co-occurrence. Bookshelves and a window appear in different panels of the same scene, supporting answer D.}
\label{fig:question_cooccurrence}
\end{figure}

Figures~\ref{fig:question_action} and~\ref{fig:question_cooccurrence} illustrate two memory-grounded questions produced by the Questioner. The displayed memory notes identify the records supporting each reference answer.

\paragraph{Action recognition (perception).}
Figure~\ref{fig:question_action} asks which craft action is performed. The micro-event at 2132--2138 seconds records a person spraying textured paint onto a pink plastic plate, supporting option A. The question tests recognition of a visible action, with its reference grounded in the event record rather than an interpretation of the subsequent painting steps.

\paragraph{Object reasoning (entity co-occurrence).}
Figure~\ref{fig:question_cooccurrence} asks which background element appears alongside the bookshelves. The split-screen scene contains bookshelves behind the participant in the bottom panel and a window in the top-left panel, supporting option D. This question belongs to the entity-co-occurrence form of object reasoning: the answer is grounded in the entities recorded within the same segment, not physical adjacency in one room.

\subsection{Solver Harness Evolution}
\label{app:solver_case}

\begin{figure}[htbp]
\centering
\includegraphics[width=\linewidth]{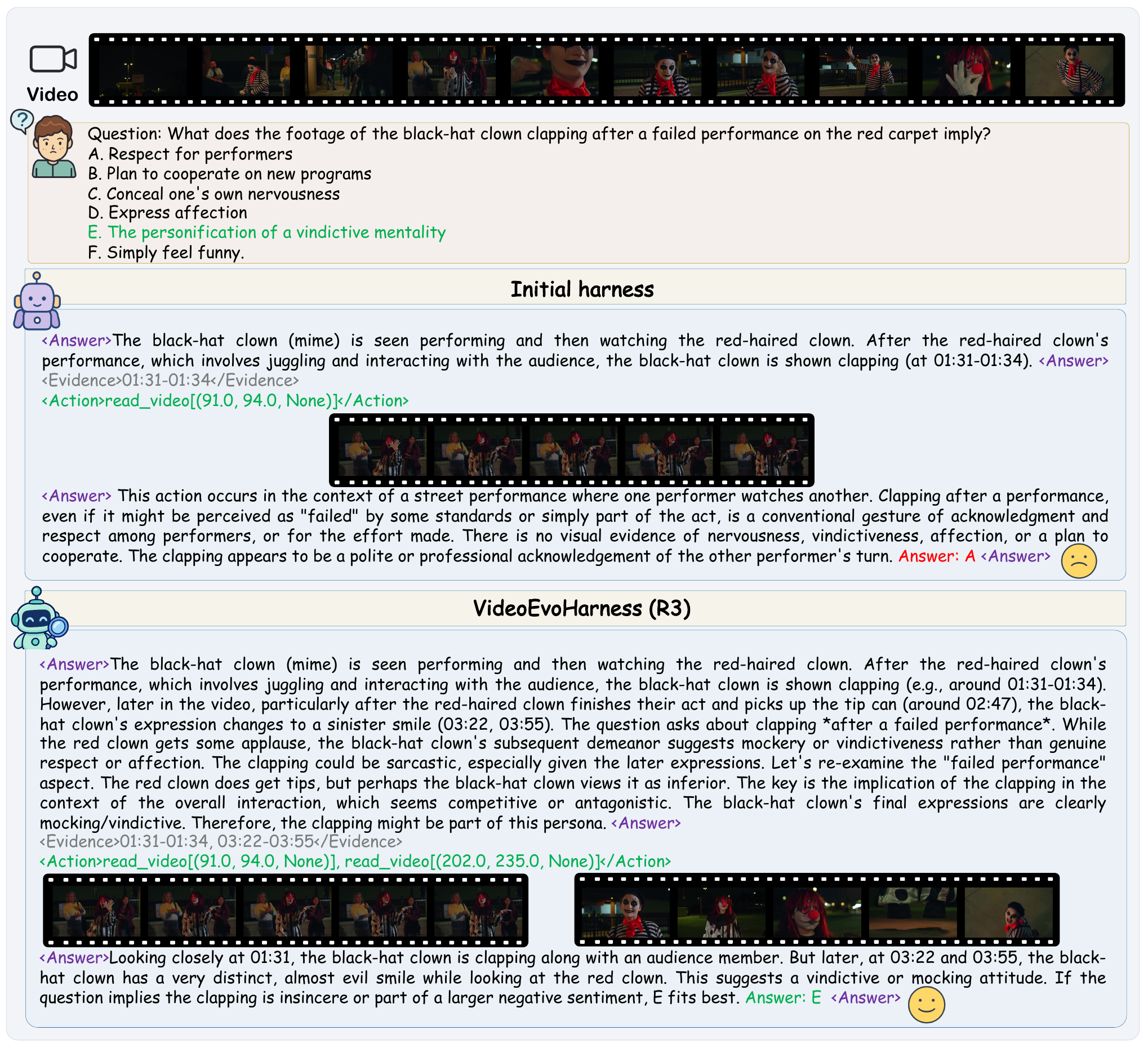}
\caption{Comparison of the initial and R3 Solver harnesses on Video-Holmes with Qwen3.7 Plus. Compared with the initial harness, the R3 harness supplements inspection of the clapping interval with later evidence, leading to answer E rather than A. The comparison illustrates a change in evidence acquisition while the backbone remains frozen.}
\label{fig:case_solver}
\end{figure}

Figure~\ref{fig:case_solver} compares the initial and R3 harnesses on a Video-Holmes question using Qwen3.7 Plus with frozen model parameters. The question asks what a black-hat clown's clapping after another performer's failed act implies. The initial harness targets the clapping interval at 01:31--01:34 for frame inspection. It interprets the gesture as conventional acknowledgment between performers and selects A, ``Respect for performers.'' This interpretation focuses on the local action without establishing its meaning in the broader interaction.

The R3 harness inspects the same interval and additionally requests frames from 03:22--03:55. These later observations provide context about the clown's attitude toward the other performer. The Solver interprets the subsequent expressions as mocking rather than respectful and selects E, ``The personification of a vindictive mentality.'' By connecting the focal gesture with evidence from a later interval, the evolved harness supports a more context-sensitive answer. This case illustrates how harness evolution improves answering through broader temporal evidence acquisition, rather than through changes to the question or model parameters.

\subsection{Strategy and Hyperparameter Evolution}
\label{app:strategy_parameter_cases}

\begin{figure}[H]
\centering
\includegraphics[width=\linewidth,height=0.64\textheight,keepaspectratio]{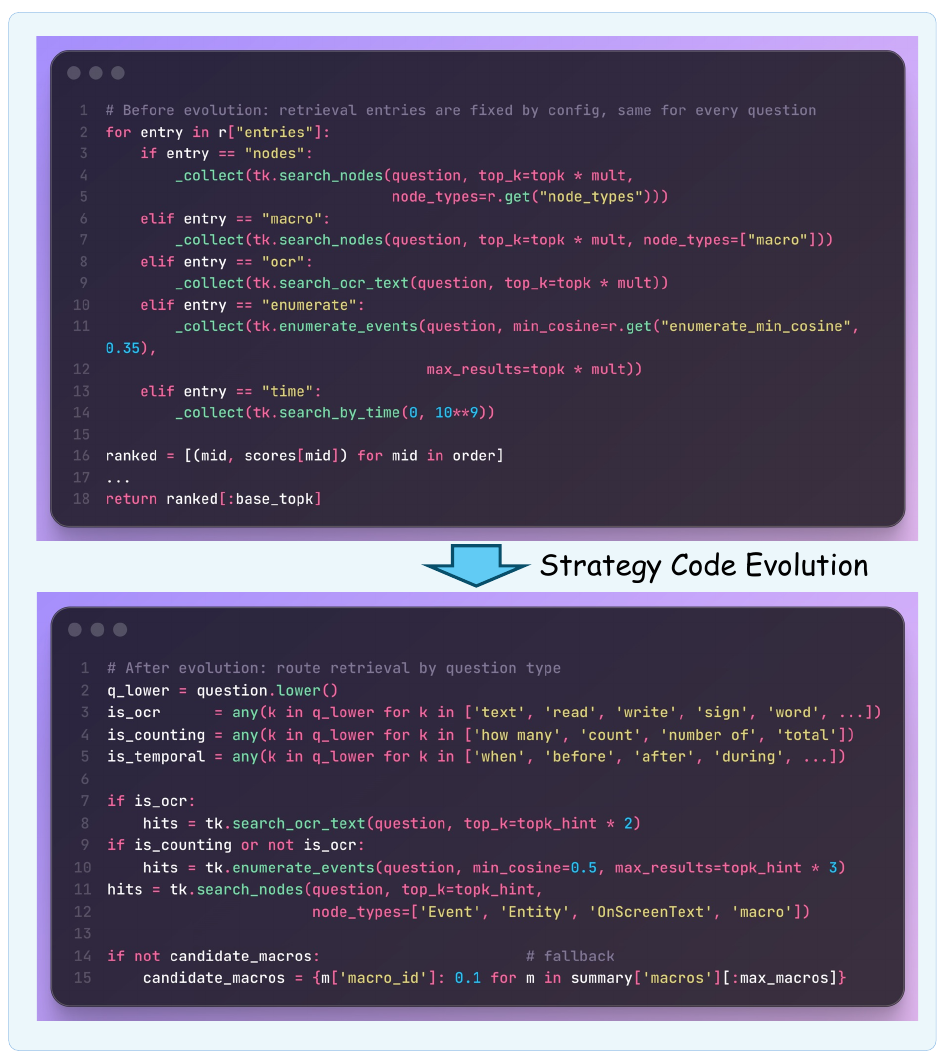}
\caption{Illustrative strategy code change during evolution. The generated routine combines keyword-conditioned retrieval with semantic search and a summary-based fallback.}
\label{fig:strategy_code_case}
\end{figure}

\begin{figure}[H]
\centering
\includegraphics[width=\linewidth]{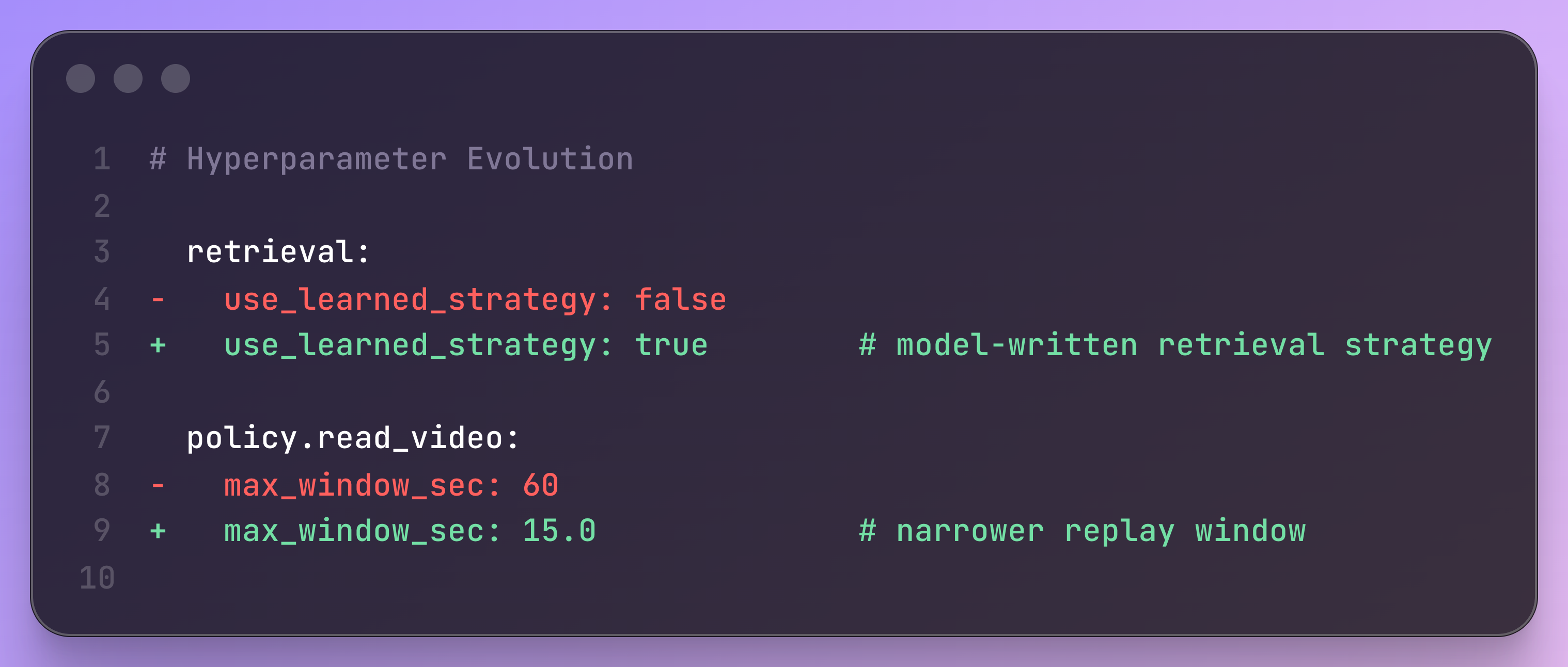}
\caption{Illustrative configuration changes during evolution: enabling the generated retrieval strategy and reducing the replay-window cap from 60 to 15 seconds.}
\label{fig:hyperparameter_case}
\end{figure}

Figures~\ref{fig:strategy_code_case} and~\ref{fig:hyperparameter_case} illustrate changes to retrieval logic and evidence allocation during evolution.

\paragraph{Strategy code.}
The original procedure in Figure~\ref{fig:strategy_code_case} uses a configured set of retrieval entry points for every question. The generated strategy instead uses question keywords to activate OCR search for text-related queries and event enumeration for counting or non-OCR queries, while retaining semantic node search. It ranks the retrieved segments and assembles their records according to the question: visible text is included for OCR queries, and event descriptions for counting and temporal queries. If retrieval yields no candidates, a summary-based fallback supplies candidate segments. This example shows how code evolution introduces question-dependent tool use and evidence assembly, rather than simply increasing the number of retrieved records.

\paragraph{Hyperparameters.}
Figure~\ref{fig:hyperparameter_case} shows two configuration changes: \texttt{use\_learned\_strategy} switches from \texttt{false} to \texttt{true}, enabling the generated retrieval routine, and \texttt{max\_window\_sec} decreases from 60 to 15 seconds. The latter limits the duration of each replay window on the memory path, restricting local inspection to a shorter interval rather than increasing the frame budget. Together, these examples illustrate how evolution adapts both the evidence-gathering procedure and its temporal scope within a frozen-backbone harness.

\clearpage

\section{Limitations and Future Plans}
\label{app:limitations}

Despite the gains observed across benchmarks and backbones, VideoEvoHarness has several limitations that motivate further research. First, improvements are not uniform across capabilities. Some categories decline as evolution progresses, even when overall accuracy increases. Future work could explore capability-aware validation and update selection to address weaker abilities while preserving existing strengths.

Second, although the framework supports evolution of skills, strategies, hyperparameters, and tools, the accepted changes in our current experiments do not cover all four components. Under the three-round budget, the retained Solver updates concern strategies and hyperparameters, while skills and tools remain unchanged. The cost of candidate generation and evaluation limits our ability to extend evolution and systematically characterize its longer-term behavior. Longer runs under controlled budgets, together with more efficient candidate evaluation, could clarify when different components change and how their contributions interact.

Finally, our evaluation relies on conventional video understanding benchmarks. These measure downstream task performance but are not designed to assess the evolution process itself. Dedicated benchmarks and protocols could evaluate sustained improvement, capability retention, generalization to changing tasks, and the cost of acquiring these gains. Such evaluation would provide a more direct account of an agent's ability to evolve beyond endpoint accuracy.

Taken together, our experiments support self evolving video agent harnesses as a promising research direction. More balanced capability development, broader exploration of component evolution, and dedicated evaluation offer concrete opportunities to advance this direction.

\end{document}